\documentclass[runningheads]{llncs}
\usepackage{color}
\usepackage{graphicx}
\newcommand{\amb}[2]{{\tt #1}$\updownarrow${\tt #2}}
\newcommand{\letter}[1]{`{\tt #1}'}
\newcommand{\letterc}[1]{`{\tt #1},'}
\newcommand{\letterp}[1]{`{\tt #1}.'}
\newcommand{\ambplain}[2]{{#1}$\updownarrow${#2}}

\usepackage{amsmath,amssymb}
\usepackage{url}
\usepackage{algorithmic}
\usepackage{algorithm}
\usepackage{here}

\begin{document}
\title{Ambigram Generation by A Diffusion Model}%



\author{Takahiro Shirakawa\inst{1}\and
Seiichi Uchida\inst{1}\orcidID{0000-0001-8592-7566}}
\authorrunning{T. Shirakawa et al.}
\institute{Kyushu University, Fukuoka, Japan\\
\email{uchida@ait.kyushu-u.ac.jp}
}

\maketitle              
\vspace{-5mm}
\begin{abstract}
{\em Ambigrams} are graphical letter designs that can be read not only from the original direction but also from a rotated direction (especially with 180 degrees). 
Designing ambigrams is difficult even for human experts because keeping their dual readability from both directions is often difficult. 
This paper proposes an ambigram generation model.  As its generation module, we use a diffusion model, which has recently been used to generate high-quality photographic images.  By specifying a pair of letter classes, such as `A' and `B,' the proposed model generates various ambigram images which can be read as `A' from the original direction and `B' from a direction rotated 180 degrees. Quantitative and qualitative analyses of experimental results show that the proposed model can generate high-quality and diverse ambigrams.
In addition, we define {\em ambigramability}, an objective measure of how easy it is to generate ambigrams for each letter pair. For example, the pair of `A' and `V' shows a high ambigramability (that is, it is easy to generate their ambigrams), and the pair of `D' and `K' shows a lower ambigramability. The ambigramability gives various hints 
of the ambigram generation not only for computers but also for human experts.
The code can be found at \url{https://github.com/univ-esuty/ambifusion}.
\end{abstract}

\section{Introduction\label{sec:intro}}
\begin{figure}[t]
\centering
\includegraphics[width=0.80\textwidth]{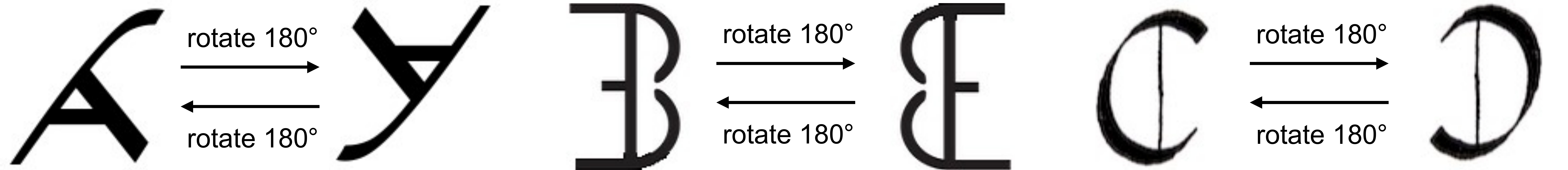}\\[-3mm]
\caption{Examples of ambigram letters designed by human experts. According to our notation, they are \amb{A}{Y},  \amb{B}{E}, and \amb{C}{D}, respectively.  These ambigrams are provided by the website {\it MakeAmbigrams} ({\tt https://makeambigrams.com/}).}
\label{fig:ambigram}
\end{figure}

{\em Ambigrams} are graphical letter designs that can be read not only from the original direction but also from a rotated direction (especially with 180 degrees).
Fig.~\ref{fig:ambigram} shows several examples of ambigram letters designed by human experts. 
The leftmost example in Fig~\ref{fig:ambigram} is an ambigram for \letter{A} and \letterp{Y} Hereafter, we denote such an ambigram as \amb{A}{Y}. The middle and rightmost samples are \amb{B}{E} and \amb{C}{D}, respectively.


Designing ambigrams is difficult even for human experts due to the following two reasons. First, we must deform letter shapes to realize the dual readability from two directions. Moreover, the deformations for ambigrams are very different from those for general font styles. This means there is no standard way to design ambigrams. 
Second, creating their ambigrams is inherently difficult for many letter class pairs. For example, it will be almost impossible to design an ambigram \amb{D}{K}, although it is easy for \amb{A}{V}.\par

Designing ambigrams is also difficult for computers. One might think that generative adversarial networks (GANs) can generate new ambigrams by using ambigrams by human experts as ``real'' image samples. However, there is no large ``real'' ambigram dataset with sufficient style variations. Therefore, conventional font generation methodologies by GANs\cite{azadi2018multi,hayashi2019glyphgan,xie2021dg} 
cannot be used for ambigrams. Furthermore, computers will suffer from the above two difficulties, like human experts.\par

\begin{figure}[t]
\centering
\includegraphics[width=.95\textwidth]{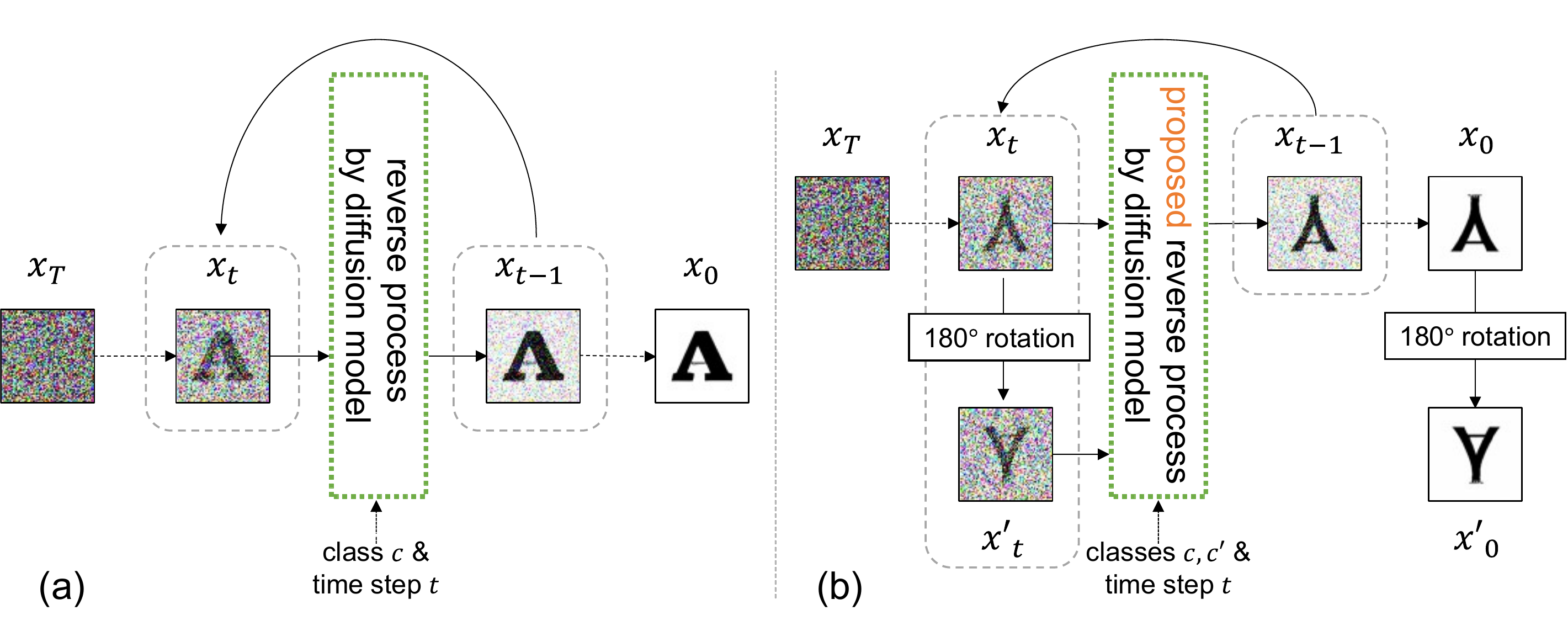}\\[-5mm]
\caption{Overview of (a)~the standard diffusion model for image generation and (b)~the proposed model for ambigram generation.} 
\label{fig:overview}
\end{figure}

This paper proposes a diffusion model for this challenging task of generating ambigrams. Fig.~\ref{fig:overview} shows an overview of (a)~the standard diffusion model for image generation and (b)~the proposed model for ambigram generation. The standard diffusion model generates a letter image $x_0$ of the class $c$ by an iterative process called {\em reverse process}. In contrast, the proposed model generates a letter image $x_0$ for class $c$ and its 180$^\circ$-rotated version $x'_0$ for class $c'$.  For example, suppose two classes $c=$\letter{A} and $c'=$\letter{Y} are specified as conditions for the original and rotated directions, respectively. Then, the diffusion model is expected to produce an image $x_0$ of \letter{A} such that the 180$^\circ$-rotated image $x'_0$ is \letterp{Y}
\par
%
An important property of the proposed model is that it does not need to have any real ambigram samples made by human experts. Instead, it needs just standard letter images of all classes. After learning from those images what shapes are readable as each letter class, the proposed model generates ambigrams with dual  readability for $c$ and $c'$. This property effectively generates a variety of unique and unexpected ambigrams, even in the current situation where reference ambigrams by human experts are not readily available enough.
\par
%
In the experiments, we first observe that the proposed model can generate high-quality and various ambigrams. In addition to this qualitative observation, we conduct a quantitative evaluation by using the recognition accuracy of a letter classifier.
If we achieve high accuracy, the generated ambigrams will have sufficient readability. The ambigrams obtained from the proposed model are then compared with those obtained from human experts and a GAN-based generator.\par
%
The above letter classification results reveal {\em ambigramability}, which is newly introduced in this paper for evaluating how easy to generate ambigrams for each class pair. 
As noted above, generating ambigrams of \amb{A}{V} will be easy, whereas \amb{D}{K} is difficult. Therefore, the former has a higher ambigramability and the latter has a lower ambigramability. It is useful to know the ambigramability when we design a word or a phrase by ambigrams. When we find a word ambigram that needs a letter pair with low ambigramability, it will be better to think of another word or phrase (or use the capital letter instead of the lower case or vice versa). The ambigramability can be automatically evaluated by recognizing  generated ambigrams by a letter classifier.\par

The main contributions of this paper are summarized as follows.
\begin{itemize}
    \item To the authors' best knowledge, it is the first attempt to generate ambigrams automatically. 
    \item We propose a new diffusion model for generating ambigrams in various styles without any real ambigram examples.
    \item We also propose ``ambigramability'' to measure the easiness of ambigram generation of each class pair objectively and automatically. 
    \item Comparative studies  reveal that ambigrams by the proposed model have more diversity while keeping their readability. 
\end{itemize}

\section{Related Work}
\begin{figure}[t]
\centering
\includegraphics[width=0.9\textwidth]{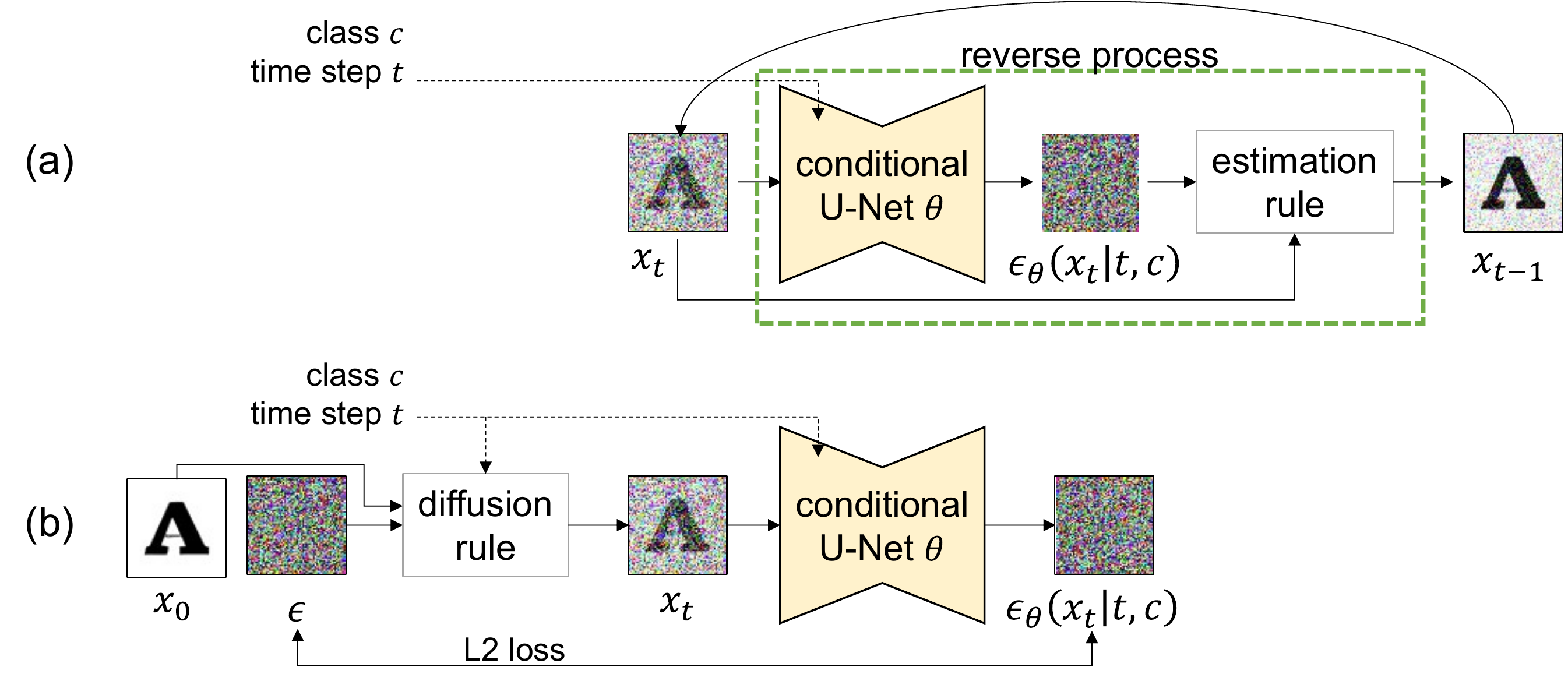}\\[-3mm]
\caption{Diffusion Denoising Probabilistic Model (DDPM). (a)~The reverse process to estimate $x_{t-1}$ from $x_t$. (b)~The training process of the conditional U-Net.} 
\label{fig:ddpm}
\end{figure}

\subsection{Diffusion Denoising Probabilistic Model (DDPM)}
Diffusion Denoising Probabilistic Model (DDPM)~\cite{diffusion-basic} is
one of the most popular diffusion models~\cite{Yang2022DiffusionMA}. DDPM assumes that a noise image $x_T$ is derived by adding a small amount of Gaussian noise to the original (clean) image $x_0$ $T$ times. (This iterative noise addition process is called {\em diffusion process}.) The highlight of DDPM is its {\em reverse process} to recover $x_0$ from $x_T$ by iterating noise removal $T$ times. Fig.~\ref{fig:ddpm}~(a) shows the reverse process. A conditional U-Net (with the weight parameter set denoted by $\theta$) estimates the noise $\epsilon_\theta(x_t \mid t, c)$ of $x_t$  under the condition that $x_t$ is an image from class $c$ after $t$ times of noise addition. Then, $x_{t-1}$ is estimated from $x_t$ and $\epsilon_\theta(x_t \mid t, c)$ by an estimation rule\footnote{Precisely speaking, the following equation~\cite{diffusion-basic} is used as the estimation rule in Fig.~\ref{fig:ddpm}~(a):
$$
x_{t-1} = \frac{1}{\sqrt{\alpha_t}}\left(x_t-\frac{1-\alpha_t}{\sqrt{1-\bar{\alpha}_t}}\epsilon_\theta(x_t \mid t, c)\right)+\sigma_tz, 
$$
where $\bar{\alpha}_t = \alpha_1\cdots\alpha_t$, $\alpha_t=1-\beta_t$, and $\beta_t$ is a constant defined by $(\beta_T - \beta_1)(t-1)/(T-1)+\beta_1$ with hyperparameters $\beta_1$ and $\beta_T$. 
$\sigma_t$ is also a time-dependent
constant determined by $\beta_t$ and $z$ is a random vector from $\mathcal{N}(0,\mathbf{I})$.
For more details, please refer to the original DDPM paper~\cite{diffusion-basic}.\label{footnote:reverse-process}}.  
By iterating this reverse process from $t=T$ to $1$, we have $x_0$ as a generated image. 
Note that the generated image $x_0$ varies even from the same $x_T$ because of the stochastic behavior of the estimation rule.  
 \par
Training DDPM is equivalent to training the conditional U-Net $\theta$. As shown in Fig.~\ref{fig:ddpm}~(b), the U-Net is trained to output $\epsilon_\theta(x_t \mid t, c)\sim \epsilon$, where $\epsilon\sim \mathcal{N}(0,\mathbf{I})$ is the noise to prepare $x_t$ from $x_0$ according to the diffusion rule defined in \cite{diffusion-basic}. The conditions $t$ and $c$ are fed to the U-Net
using an embedding technique like position embedding~\cite{vaswani2017attention}.
\par
\subsection{Letter Image Generation}
For generating letter images (or font images), various image generation models, especially  GANs\cite{radford2015unsupervised,karras2017progressive,brock2018large,karras2019style,lee2021vitgan}, have been used.
The GAN-based models can also generate letter images with conditions, such as class labels\cite{mirza2014conditional,hayashi2019glyphgan,karras2020analyzing,karras2021alias} or texts\cite{xu2018attngan,Patashnik_2021_ICCV}. Recently, diffusion models have achieved high-quality photographic image generation~\cite{dalle2,ldm,imagen}.  
All the above methods cannot be used for the ambigram generation directly; this is simply because i)~they need real ambigrams as their training samples and ii)~they have no mechanism to realize the dual readability as ambigrams. Consequently, to the authors' best knowledge, there has been no past attempt to generate ambigrams by computers. Note again that our model does not need any reference ambigram examples.
\section{Ambigram Generation by A Diffusion Model} \label{sec:ambigram-generation}

\begin{figure}[t]
\centering
\includegraphics[width=0.8\textwidth]{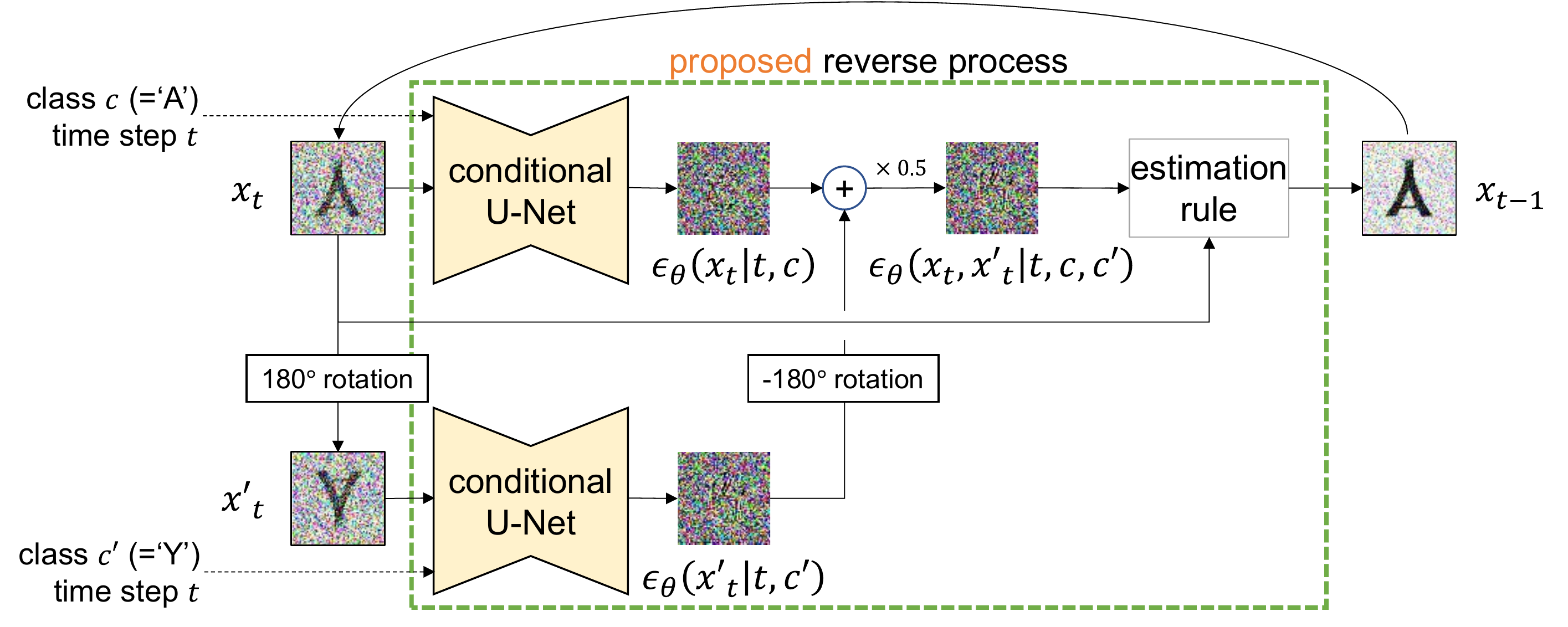}
\caption{The proposed reverse process for the ambigram generation.} 
\label{fig:proposed}
\end{figure}

\subsection{Outline} \label{sec:generation-architecture}
Fig.~\ref{fig:proposed} shows the proposed reverse process to generate ambigrams. 
Inputs to the process are a noisy image $x_t$ and its 180$^\circ$-rotated version $x'_t$. If we aim to generate ambigrams of \ambplain{$c$}{$c'$}, two class labels $c$ and $c'$ are given as conditions. Then, the reverse process will generate $x_{t-1}$ with less noise than $x_t$. We expect that $x_{t-1}$ becomes more readable as class $c$, whereas its 180$^\circ$-rotated version (i.e., $x'_{t-1}$) does $c'$. Preparing a noise image $x_T$ and then repeating this reverse process until $t=0$, we will have noise-free $x_0$ and $x'_0$ (as shown in the rightmost part of Fig.~\ref{fig:overview}(b)) that form an ambigram \ambplain{$c$}{$c'$}. \par 
The main difference between the proposed model of Fig.~\ref{fig:proposed} and the standard DDPM of Fig.~\ref{fig:ddpm}~(a) is that the former estimates two noise images $\epsilon_\theta(x_t \mid t, c)$ and $\epsilon_\theta(x'_t \mid t, c')$ by applying the same U-Net $\theta$ to $x_t$ and $x'_t$. Then, these noise images are simply averaged to have a single noise image $\epsilon_\theta(x_t, x'_t \mid t, c, c')$: 
\begin{equation}
\epsilon_\theta(x_t, x'_t \mid t, c, c') =\left(\epsilon_\theta(x_t \mid t, c) + \operatornamewithlimits{Rot}_{-180^\circ}\left[\epsilon_\theta(x'_t \mid t, c')\right]\right)/2.
\end{equation}
Note that $\epsilon_\theta(x'_t \mid t, c')$ is rotated with $-180^\circ$ (by the operator $\operatornamewithlimits{Rot}[\ ]$) before taking average, for having the original direction as $\epsilon_\theta(x_t \mid t, c)$. The averaged noise image is useful to generate $x_{t-1}$ with an appearance of $c$ and (rotated) $c'$, that is, useful to generate an ambigram \ambplain{$c$}{$c'$}.  \par 
%
Training the proposed model of Fig.~\ref{fig:proposed} is rather straightforward. We use the training process of the standard DDPM, shown in Fig.~\ref{fig:ddpm}(b), for the proposed model. Only the difference from Fig.~\ref{fig:ddpm}(b) is that the proposed model uses the averaged noise image $\epsilon_\theta(x_t, x'_t \mid t, c, c')$ instead of $\epsilon_\theta(x_t \mid t, c)$. The averaged noise image guides $x_{t-1}$ to the intermediate direction between $c$ and $c'$. Before this training, we also perform a pretraining step, as will explain below. As already emphasized, we use ordinary font images as $x_0$ and do not use any ambigram examples.
\subsection{Details} \label{sec:additional-techs}
\subsubsection{Pretraining for Ordinary Letter Image Generation}
Before generating ambigrams by the model of Fig.~\ref{fig:proposed},  its conditional U-Net $\theta$ is pretrained to generate ordinary (i.e., non-ambigram) letter images. In other words, we first train the U-Net $\theta$ in the standard DDPM scheme of Fig.~\ref{fig:ddpm}~(b). We also use an ordinary font dataset for pretraining. The pretrained U-Net can produce various ordinary fonts with the reverse process of Fig.~\ref{fig:ddpm}~(a).
\subsubsection{Classifier-free Guidance}
In the reverse process, we employ {\em classifier-free guidance}~\cite{classifier-free}. In our ambigram generation task, this technique can control the trade-off between the readability and diversity of the generated ambigrams. By setting its {\em guidance strength} parameter $s$ larger, the generated ambigrams become more readable with less diversity. In the later experiment, we will observe the effect of the parameter value $s$. 
\par
\subsubsection{Re-spacing Technique}
{\em Re-spacing}~\cite{diffusion-improved,diffusion-beating} is a technique to accelerate the reverse process by skipping some time steps $t$. 
In general, diffusion models (including DDPM) need to set $T$ at a large value (say, 1,000) for appropriate results. This means we must repeat the reverse process 1,000 times to generate just one image. The re-spacing technique allows $t$ to decrease with $\Delta$ interval, and therefore the total repetition time is reduced from $T$ to $T/\Delta$. In \cite{diffusion-beating}, it is reported that this technique keeps the quality of generated images.
\subsubsection{Data Augmentation by Horizontal Shift} \label{sec:da-ambigram}
A simple but effective technique to generate more readable and diverse ambigrams is data augmentation by horizontal shift. The augmented image of $x_0$ is just a horizontally-shifted version of $x_0$. Roughly speaking, without this technique, ambigrams \amb{E}{P} will become like 
\hbox{`{\tt E}\hspace{-4pt}\rotatebox[origin=c]{180}{{\tt P}},'} that is, `{\tt E}' and `\rotatebox[origin=c]{180}{{\tt P}}' will completely overlap. In contrast, with the technique, we will have an ambigram like `\hbox{\rotatebox[origin=c]{180}{{\tt P}}\hspace{-2pt}{\tt E}.'} Such designs are highly readable and common in ambigram designs by human experts. 
In the experiment, we will observe the actual effect of the data augmentation.
\section{Evaluation of Ambigrams and Ambigramability} \label{sec:ambigram-evaluation}
\subsection{Evaluating Ambigrams by A Classifier} \label{sec:classifier}
For evaluating the generated ambigram $x_0$ of \ambplain{$c$}{$c'$}, we will check the readability of $x_0$ as $c$, as well as that of $x'_0$ as $c'$.  
For example, we can treat an ambigram $x_0$ of \amb{C}{D} as readable when $x_0$ and its 180$^\circ$-rotated version $x'_0$ are readable as \letter{C} and \letterc{D} respectively. This readability is objectively and automatically evaluated by a conditional binary classifier $\mathcal{C}(x, c)\in \{\mathtt{True}, \mathtt{False}\}$. 
The classifier returns $\mathtt{True}$ when the input image $x$ is acceptable as the input class label $c$ and returns $\mathtt{False}$ otherwise.
Consequently, if $\mathcal{C}(x_0, c)=\mathtt{True}$ and $\mathcal{C}(x'_0, c')=\mathtt{True}$, it is a {\em successful} ambigram of \ambplain{$c$}{$c'$} with dual readability. \par
The classifier is an MLP whose input is a concatenation of two vectors: one from a standard CNN~\cite{vgg} for the image input $x_0$ and the other from a fixed random vector pre-assigned to the class $c$. The CNN for $x_0$ is pre-trained with the standard letter image classification task, and the final layer is discarded to use the CNN as a feature extractor.
\par
\subsection{Ambigramability}
One of the highlights of this paper is to introduce {\em ambigramability}, which evaluates how each alphabet letter pair is easy to generate ambigrams. As noted in Section~\ref{sec:intro}, ambigramability heavily depends on class pairs; for example, \amb{A}{V} will have a higher ambigramability, and \amb{D}{K} have a lower ambigramability. Determining the pair-wise ambigramability automatically and objectively will help further ambigram designs. 
\par
Ambigramability for the pair of $c$ and $c'$ is calculated by a sufficient number $N$ of ambigrams \ambplain{$c$}{$c'$} and the classifier in Section~\ref{sec:classifier}. Specifically, ambigramability is defined as the number of successful ambigrams among $N$ ambigrams. For simplicity, we use $N=100$ in the later experiment. Note that the ambigramability values for \ambplain{$c$}{$c'$} and \ambplain{$c'$}{$c$} will be similar but not the same due to the stochastic behavior of the diffusion model.

\section{Experimental Results \label{sec:result}} 
\subsection{Experiment Setup} \label{sec:experiment-setup}
\subsubsection{Dataset} \label{sec:experiment-setup-dataset}
As described in Section~\ref{sec:ambigram-generation}, it is enough to prepare a standard (i.e., non-ambigram) font image dataset to train the proposed model. In the following experiment, we use 52 letter images of Latin alphabet classes
(\letterc{A}$\cdots$,\letterc{Z}\letterc{a}$\cdots$,\letter{z}) from MyFonts dataset~\cite{myfonts}.
The MyFonts dataset contains about 20,000 fonts, and we remove illegible, collapsed, or excessively decorated fonts before training. Consequently, 11,991 fonts were used to train the model. Each image is binary and $64\times 64$ pixels.
We apply the data augmentation to all 52 classes of alphabets (A-z, a-z) while using 40\% of training data. The amount of horizontal shifts is randomly determined within 25\% of the image size.
\subsubsection{Implementation Details}
The model of the conditional U-Net is similar to a model used in \cite{diffusion-beating}. This U-Net is more elaborated from the original U-Net~\cite{org-unet} and consists of a stack of residual layers with an attention mechanism. We use an adaptive group normalization\cite{diffusion-beating} for embedding time step $t$ and class conditions $c, c'$.
The model was trained end-to-end with 500K iterations and batch size 64. As an optimizer, Adam\cite{adam,adamw} is used with its learning rate $10^{-4}$.
As the parameters specific to DDPM, we use $T=1,000$, $\beta_1=10^{-4}$, and $\beta_T=0.02$ for the pretraining step and the ambigram generation step.
We also set $\Delta=20$ for re-spacing and $s=5.0$ for classifier-free guidance.  
A preliminary experiment sets these parameters.\par

\begin{figure}[t]
\centering
\includegraphics[width=.8\textwidth]{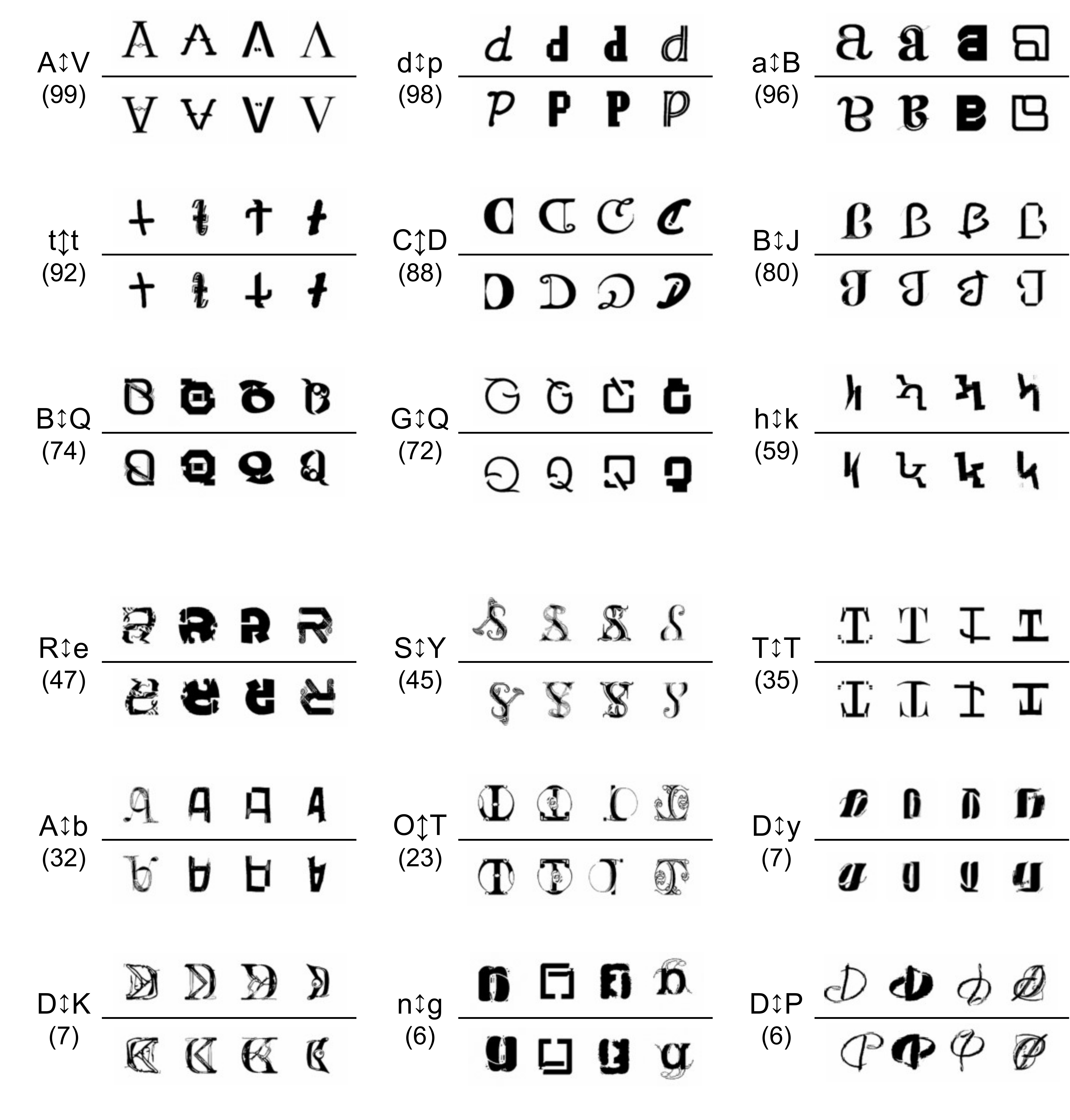}\\[-3mm]
\caption{Ambigrams generated by the proposed model. The parenthesized number is the ambigramability score ($\uparrow$) of the letter pair. The upper three rows are rather easy class pairs (with higher ambigramability scores), and the lower three are not. Note that all these ambigrams are successful ones.}  \label{fig:05-00}
\end{figure}
\subsubsection{Evaluation Metrics}
Using the trained classifier $\mathcal{C}(x,c)$ and $N=100$ generated ambigrams for each class pair, the ambigramability score ($\uparrow$) is calculated for the quality evaluation. The binary classifier $\mathcal{C}(x,c)$  is trained by the MyFonts dataset. After training, its test accuracy was 98.70\%. Similar to the past attempts to generate letter images by GANs, we also use Frechet Inception Distance (FID$\downarrow$)~\cite{heusel2017gans} for evaluating the quality and diversity of the generated ambigrams. The FID score of the generated ambigrams is calculated by comparing their feature distribution with that of the MyFonts dataset on the Inception-V3\cite{szegedy2016rethinking}.
\subsection{Qualitative Evaluations\label{sec:qualitative}}
\subsubsection{Examples of Generated Ambigrams}
Fig.~\ref{fig:05-00} shows the ambigrams generated by the proposed model. Four ambigram examples are shown for each class pair \ambplain{$c$}{$c'$}. The lower samples ($x'_0$) are 180$^\circ$-rotated versions of the upper samples ($x_0$). We expect that the upper samples are readable as $c$ and the lower $c'$. The parenthesized number is the ambigramability score $\in [0, N(=100)]$ of the letter pair. Note that all ambigrams in this figure were successful ones by the classifier $\mathcal{C}$.\par
The upper three rows of Fig.~\ref{fig:05-00} show the ambigrams for {\em easy} pairs with higher ambigramability scores.
These ambigrams realize their dual readability from both directions. 
For the pair \amb{A}{V}, the generated images of \letter{A} have a thin or no horizontal stroke; consequently, their 180$^\circ$-rotated versions are readable as \letterp{V} 
Similarly, for the pair \amb{C}{D}, the generated images of \letter{C} have a thin or short vertical stroke to make their 180$^\circ$-rotated versions readable as \letter{D}. For \amb{t}{t}, the horizontal stroke often comes in the middle of the ambigrams, even though such a style is not common in the non-ambigram fonts used for training.  \par
An important observation of those successful cases is that the proposed model based on a diffusion model can generate various ambigrams. For \amb{d}{p}, one of the easiest class pairs, the generated ambigrams have various styles, i.e., calligraphic, pixel art, slab-serif, and outlined. Other ambigrams, such as \amb{a}{B} and \amb{B}{Q}, also have wide variations in their styles. The stochastic property of the diffusion model helps to have those variations.\par
%
The lower three rows of  Fig.~\ref{fig:05-00} show the ambigrams for rather difficult class pairs (with lower ambigramability scores). Although these ambigrams were still successful ones according to the computer classifier $\mathcal{C}$, they have low readability for humans. For example, \letter{e}s by \amb{R}{e} are difficult to be read as \letterp{e}\par 
On the other hand, it is also true that the generated ambigrams give various inspirations to humans, even when they have low ambigramability scores. For example, the examples of \amb{S}{Y} and \amb{D}{K} show that there are still possibilities to design ambigrams for these pairs by using heavy but careful deformations. The ambigrams \amb{T}{T} are also inspiring; our model discovers that having a short serifed horizontal stroke and a long straight horizontal stroke can realize the dual readability of \amb{T}{T}. \par
%
\begin{figure}[t]
\centering
\includegraphics[width=0.8\textwidth]{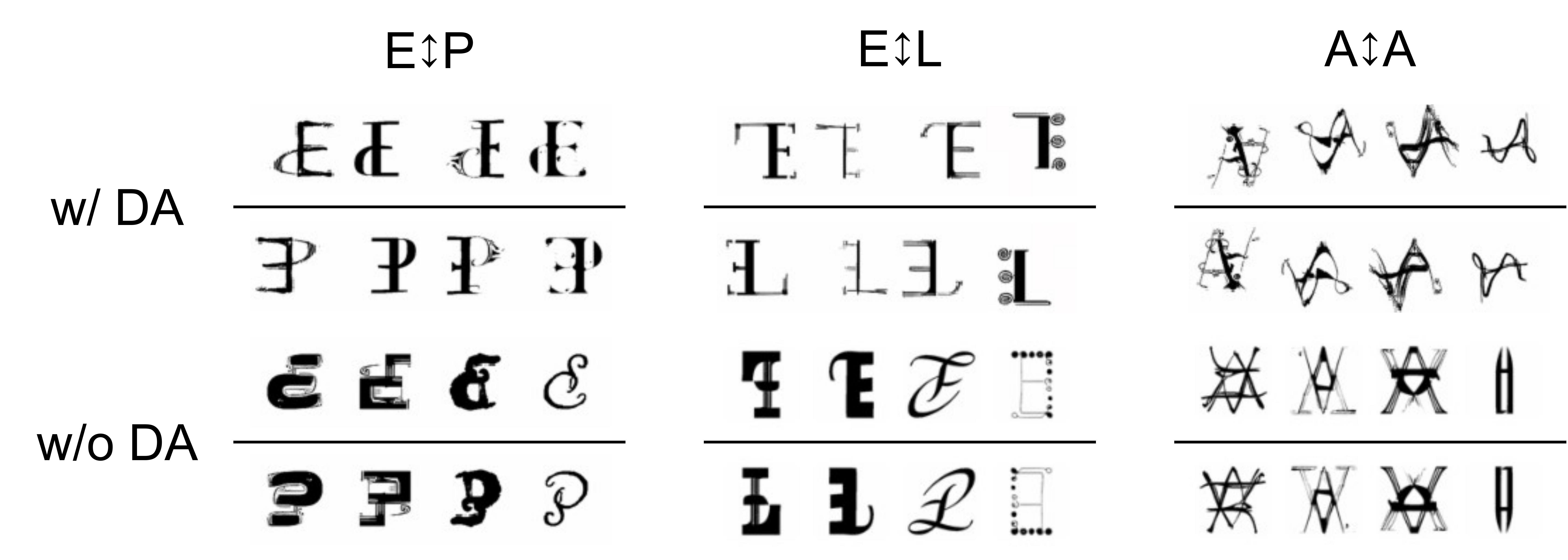}\\[-3mm]
\caption{Effect of data augmentation by horizontal shift.} \label{fig:03-00}
\end{figure}

\subsubsection{Effect of Data Augmentation by Horizontal Shift}
Fig.~\ref{fig:03-00} shows the ablation study for the data augmentation by horizontal shift. As expected, this data augmentation increases the degree of freedom in ambigram design. For example, the generated ambigrams of \amb{E}{P} suggest that the data augmentation is effective in having more readable ambigrams. (Later in Fig~\ref{fig:compare-with-manmade}, we will observe that the ambigrams of \amb{E}{P} by human experts also look like \hbox{`\rotatebox[origin=c]{180}{{\tt P}}\hspace{-2pt}{\tt E},'} instead of \hbox{`\rotatebox[origin=c]{180}{{\tt P}}\hspace{-6pt}{\tt E}.'})  Similar positive effects are found in \amb{E}{L} and \amb{A}{A}. Note that even with data augmentation, the model can generate ambigrams that would be generated without data augmentation. (The model can generate \hbox{`\rotatebox[origin=c]{180}{{\tt P}}\hspace{-6pt}{\tt E}'} if it has dual readability.)
In other words, the model automatically chooses the better one. Consequently, the data augmentation helps to increase the readability and diversity of generated ambigrams.
\begin{figure}[t]
\centering
\includegraphics[width=0.8\textwidth]{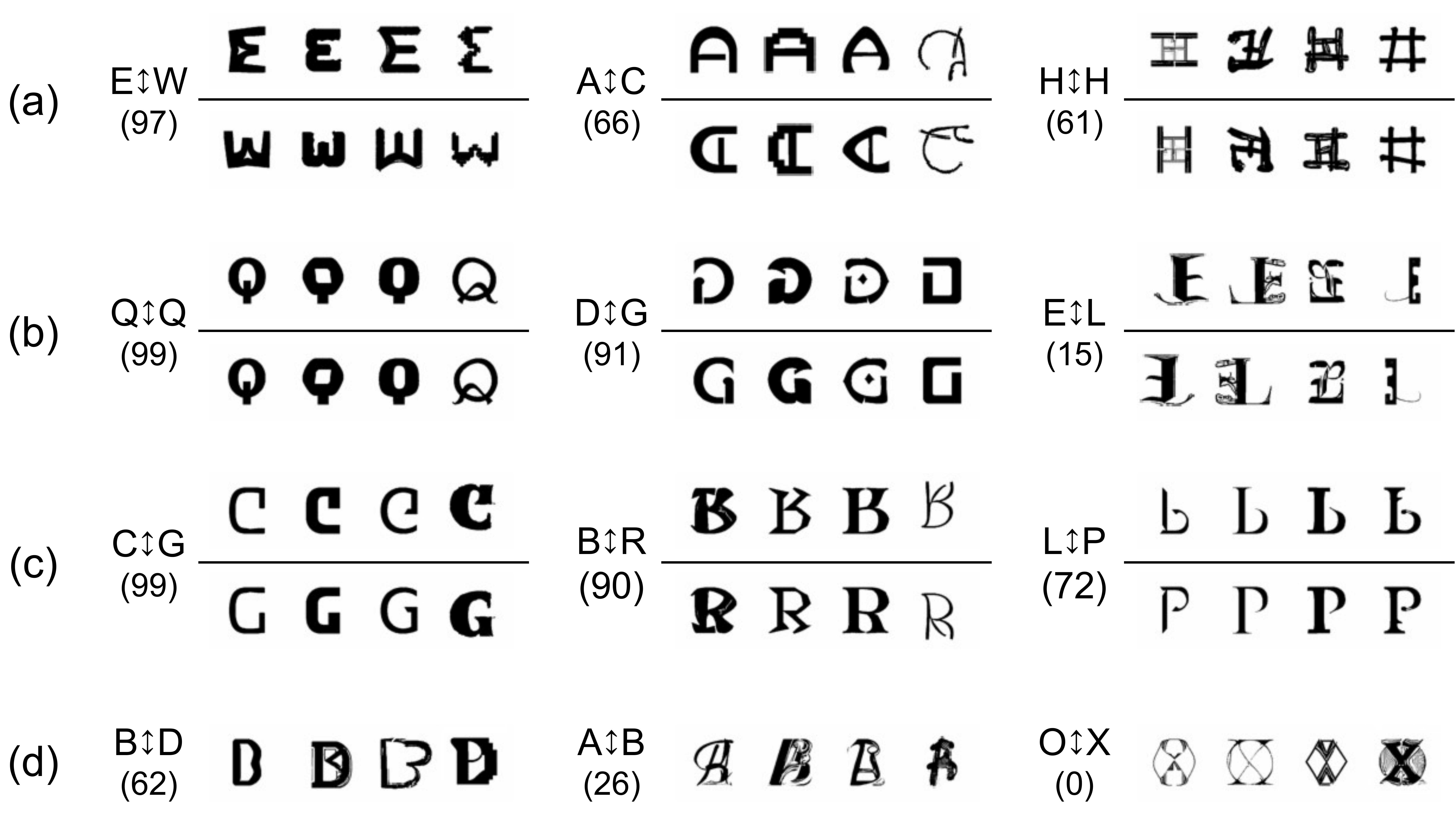}\\[-3mm]
\caption{Ambigrams for (a)~90$^\circ$-rotation, (b)~horizontal flip, (c)~vertical flip, and (d)~overlay.} \label{fig:09-00}
\end{figure}

\begin{figure}[t]
\includegraphics[width=\textwidth]{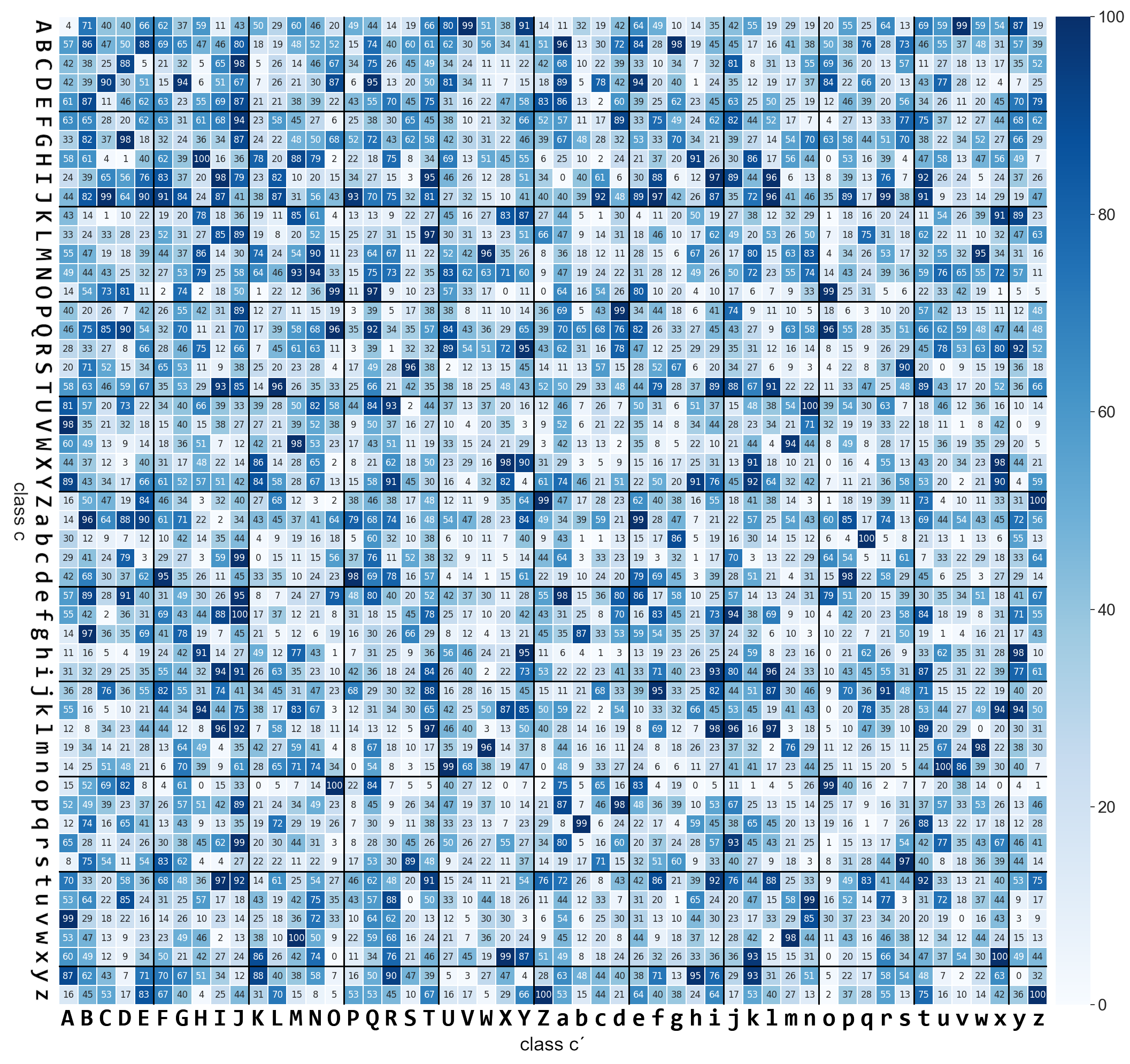}\\[-5mm]
\caption{Ambigramability score of each letter pair \ambplain{$c$}{$c'$}. A pair with a  darker cell has a higher ambigramability score; that is, generating their ambigrams is rather easy.\label{fig:06-00}}
\end{figure}
\subsubsection{Generating Different Types of Ambigrams}
Our model can generate different types of ambigrams by replacing the $\pm 180^\circ$-rotation module in Fig.~\ref{fig:proposed} with another transformation module.
Fig.~\ref{fig:09-00} shows the ambigram examples of four different types.
Ambigrams of (a), (b), and (c) are generated by $\pm 90^\circ$-rotation, horizontal flip, and vertical flip, respectively, instead of 
$\pm 180^\circ$-rotation. Consequently, they become readable as another 
letter after the transformations. Samples of Fig.~\ref{fig:09-00}~(d) are generated without any transformation --- therefore, they seem like overlaid letters (rather than ambigrams) and keep readability as two letters from their original direction.

\begin{figure}[t]
\centering
\includegraphics[width=0.7\textwidth]{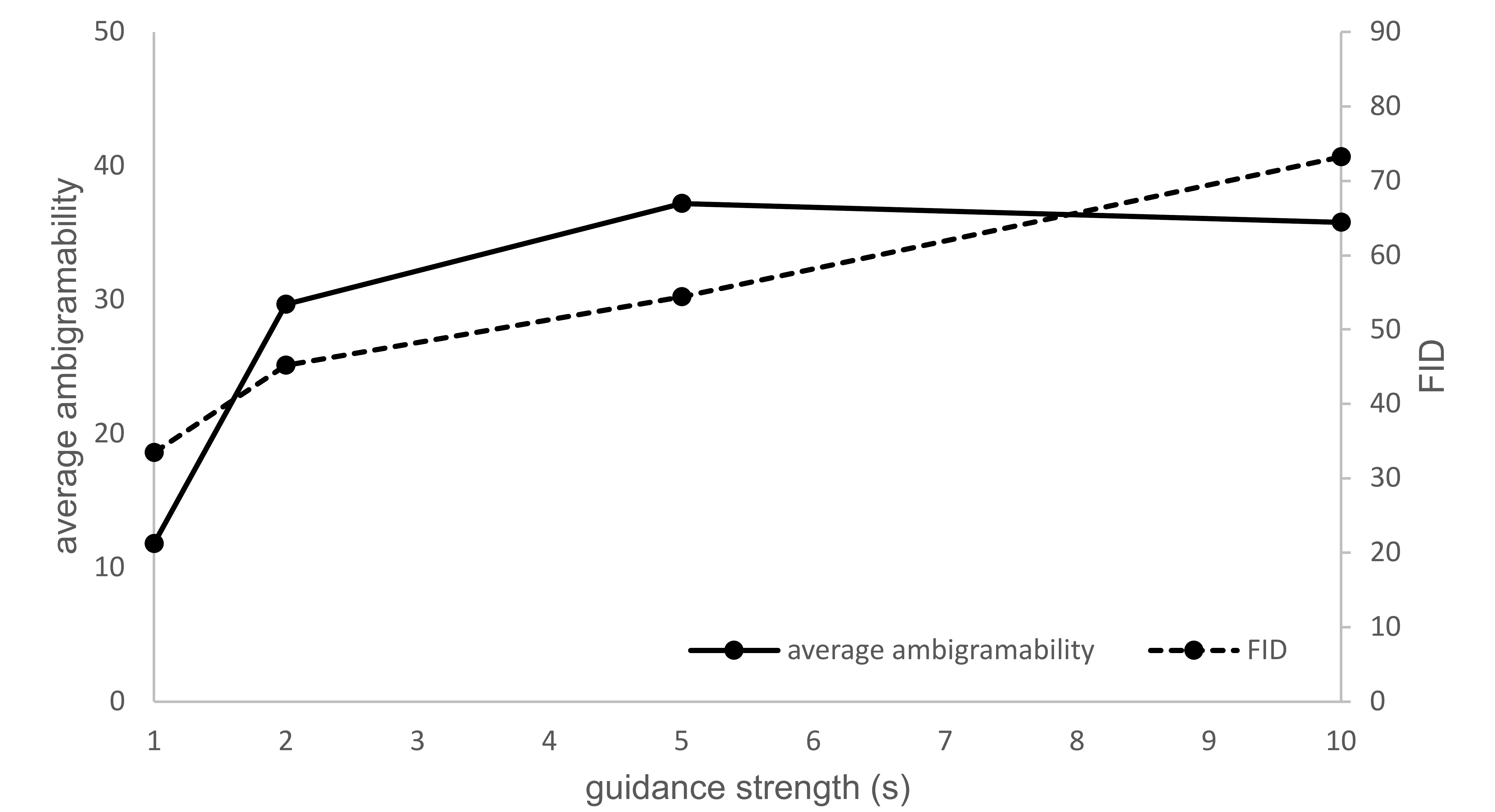}\\[-2mm] 
\caption{The trade-off between the readability and diversity of generated ambigrams by controlling the guidance strength parameter $s$.} \label{fig:02-50}
\end{figure}
\subsection{Quantitative Evaluations\label{sec:quantitative}}
\subsubsection{Ambigramability Scores}
Fig.~\ref{fig:06-00} shows the ambigramability scores for all pairs of 52 Latin alphabet letters. Since $N=100$, the score will range in $[0,100]$. If the score at the $(c,c')$-th element is high, generation of ambigrams of \ambplain{$c$}{$c'$} is easy. \par
First, Fig.~\ref{fig:06-00} proves that successful ambigrams are available for most class pairs. Almost all pairs have non-zero ambigramability, and most have more than 30. In fact, the average score is about 37. Several pairs, such as \amb{b}{q}, \amb{f}{J}, and \amb{H}{H}, have the highest score of 100. Moreover, we often find unexpectedly-high cases. For example, although generating \amb{B}{Q} and \amb{G}{Q} seems difficult, they have rather high scores of 74 and 72, respectively. These positive results suggest that our model will be useful to give inspiration to human experts for creating ambigrams.\par
We also find that several pairs, such as \amb{n}{P}, \amb{O}{X}, and \amb{y}{y}, have the lowest score of zero. Such negative information is also useful for designing  word ambigrams. For example, the zero ambigramability of  \amb{n}{P} suggests that designing a word ambigram of \amb{Pen}{Pen} is difficult and better to be avoided.
Note that Fig.~\ref{fig:09-00}~(d) shows a possible 
remedy for \amb{O}{X}; it is not an ambigram but has dual readability as \letter{O} and \letter{X} without rotation.
\par
%
In ambigrams designed by human experts, capital and small letters are used interchangeably. For example~\footnote{{\tt https://www.basilemorin.com/ambigrams.html}}, a word ambigram \amb{caNDy}{caNDy} is used instead of \amb{candy}{candy}.  This is because ambigrams \amb{a}{D} (88) and \amb{N}{N} (94) are easier than \amb{a}{d} (21) and \amb{n}{n} (44), respectively. (The parenthesized number is the corresponding ambigramability score.) Fig~\ref{fig:05-00} shows other prominent examples, such as \amb{d}{p} (98) and \amb{D}{P} (6). By referring to these differences, we can design various word-wise ambigrams with appropriately chosen capital and small letters.
\subsubsection{Trade-off between Readability and Diversity}
As described in Section~\ref{sec:additional-techs}, we introduce the classifier-free guidance to control the trade-off between readability and diversity. 
Fig.~\ref{fig:02-50} shows that the guidance strength $s$ can control the trade-off. As $s$ increases, both the average ambigramability and FID increase; this means that the readability of the generated ambigrams increases, but their diversity decreases. We, therefore, use $s=5$ in the above experiments as the best compromise for the trade-off. 
%
\begin{figure}[t]
\centering
\includegraphics[width=0.85\textwidth]{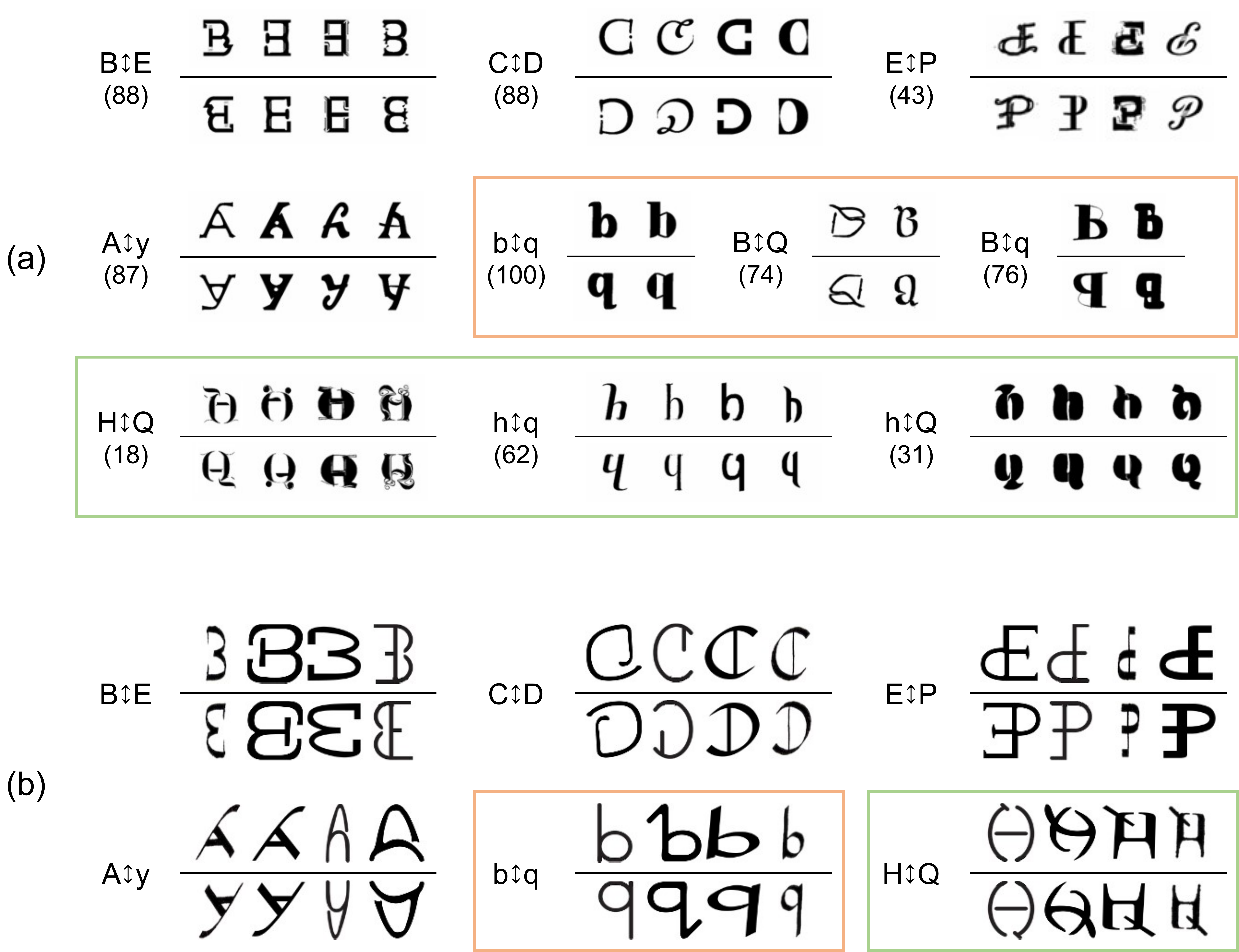}\\[-3mm]
\caption{Qualitative comparison of ambigrams by (a)~the proposed model and (b)~human experts. Since we could not find a sufficient number of ambigrams by human experts, we do not show the ambigramability scores for (b). } \label{fig:compare-with-manmade}
\end{figure}

\begin{figure}[t]
\centering
\includegraphics[width=0.55\textwidth]{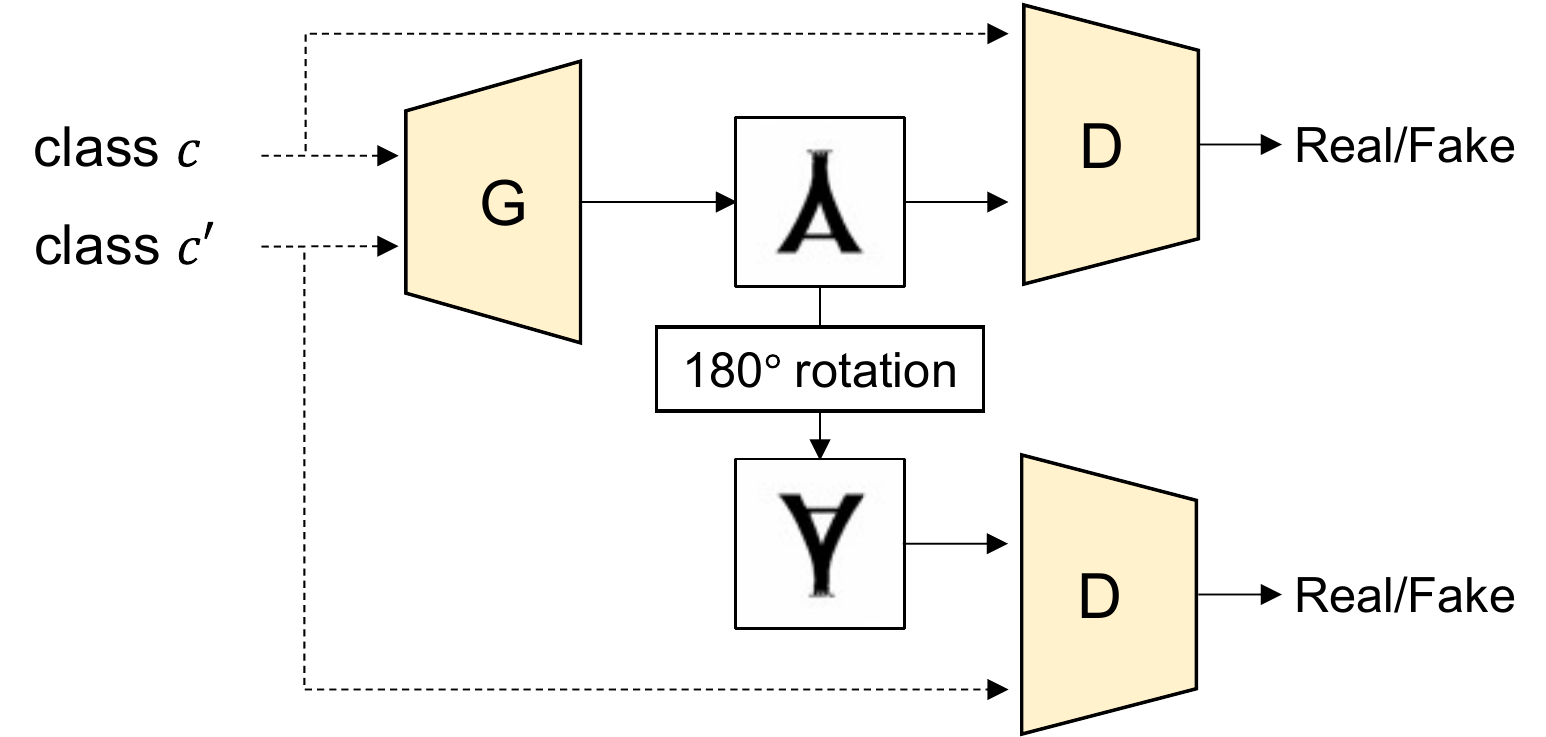}\\[-3mm]
\caption{A GAN-based ambigram generator.} \label{fig:ambigram-gan}
\end{figure}

\begin{figure}[t]
\centering
\begin{minipage}[b]{0.48\columnwidth}
\centering
\includegraphics[width=.9\columnwidth]{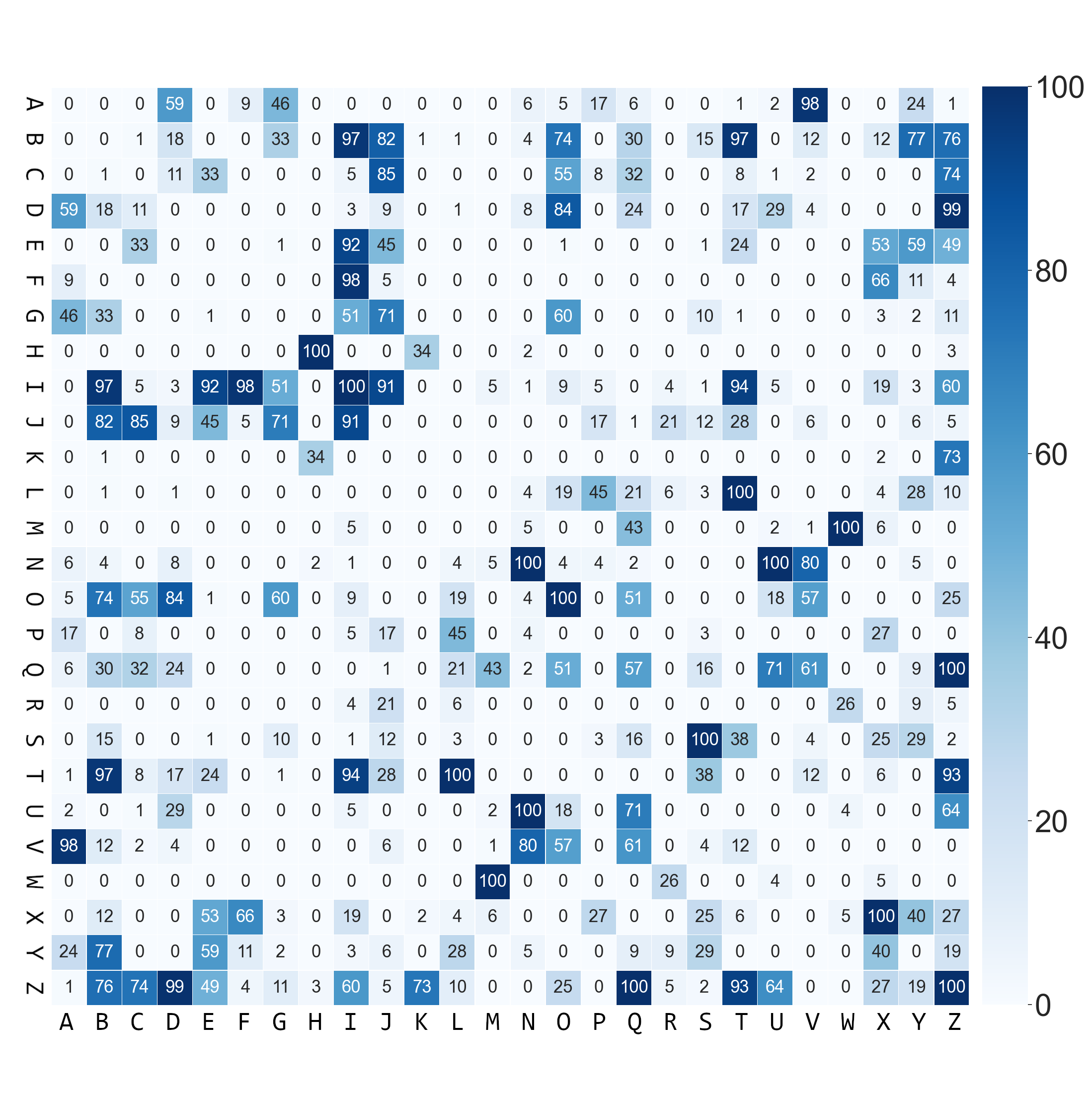}\\[-2mm]
(a)~GAN-based model.
\end{minipage}
\begin{minipage}[b]{0.48\columnwidth}
\centering
\includegraphics[width=.9\columnwidth]{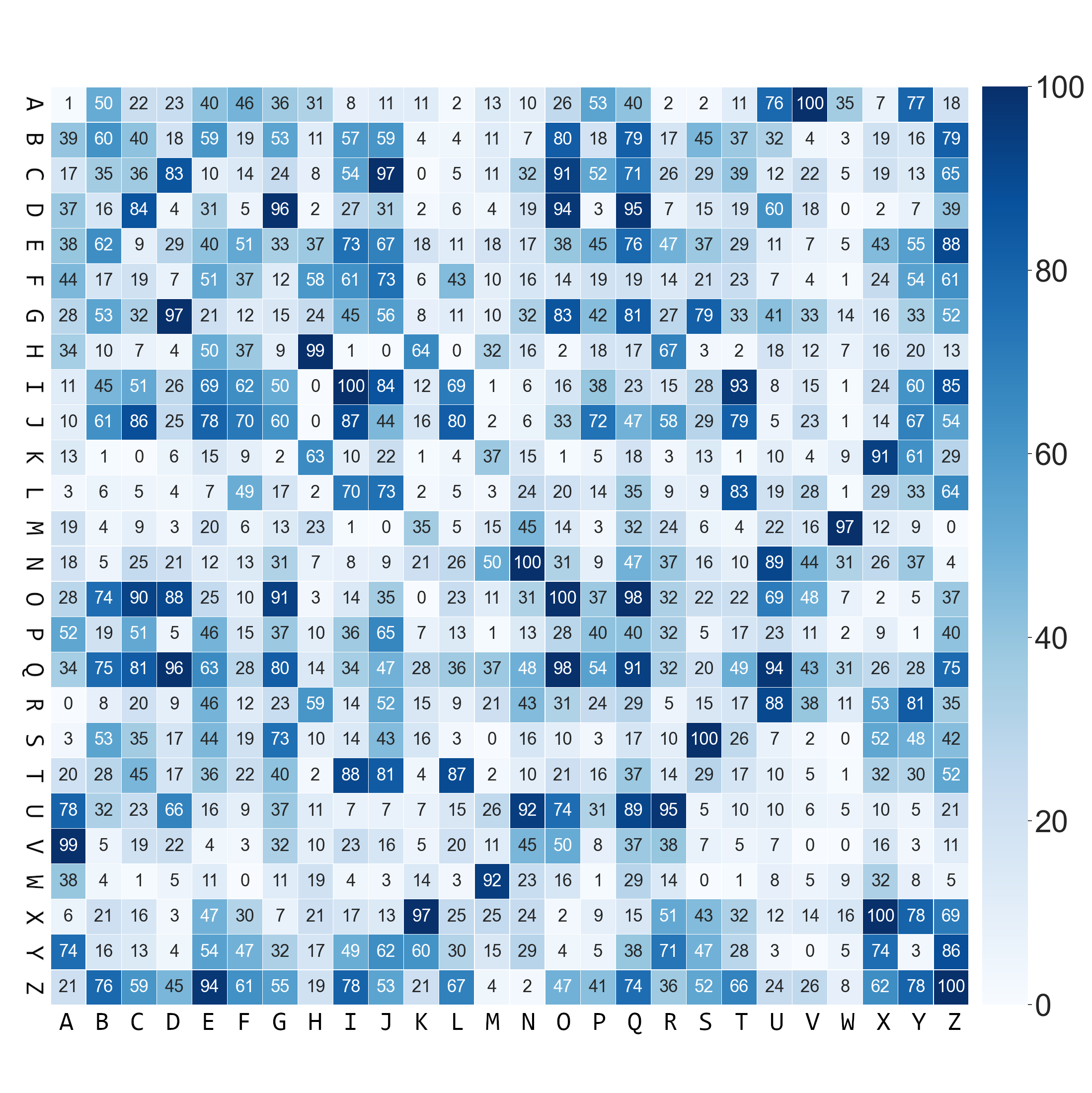}\\[-2mm]
(b)~The proposed model.
\end{minipage}
\caption{Ambigramability scores by the proposed model and the GAN-based model.} \label{fig:12-50}
\end{figure}

\begin{figure}[t]
\centering
\includegraphics[width=0.8\textwidth]{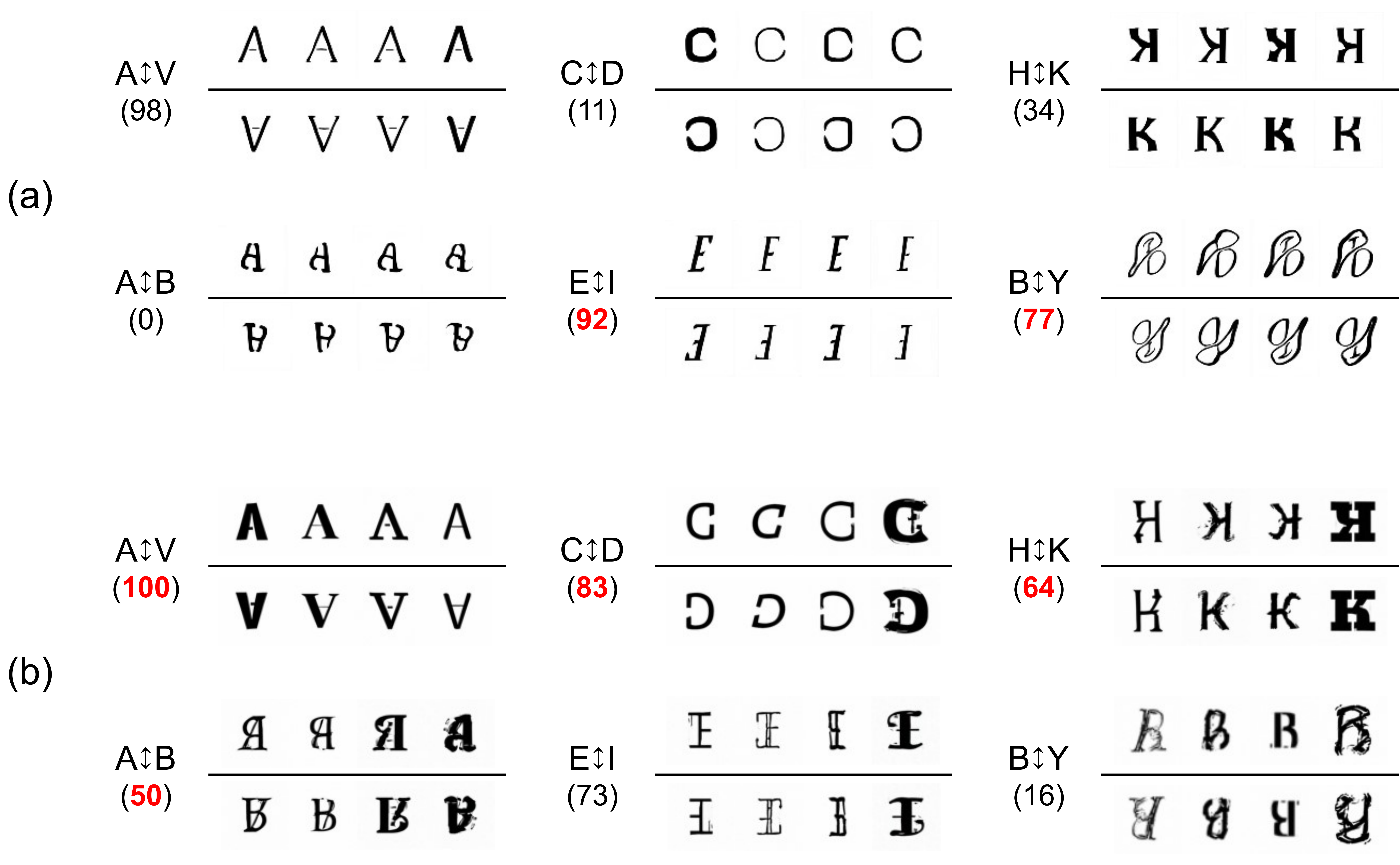}\\[-3mm]
\caption{Comparison of ambigrams by (a)~the GAN-based model and (b)~the proposed model. Ambigramability scores in \textcolor{red}{red} indicate whether (a) or (b) is the winner for a certain \ambplain{$c$}{$c'$}.} \label{fig:12-00}
\end{figure}

\subsection{Comparative Studies}
\subsubsection{Comparison with Ambigrams by Human Experts}
Fig.~\ref{fig:compare-with-manmade} compares ambigrams by (a)~the proposed model and (b)~human experts. 
The latter samples were collected from the website {\tt MakeAmbigrams}\footnote{{\tt https://makeambigrams.com/}}, where eight types of man-made ambigram fonts are provided.
\par
For \amb{B}{E}, \amb{C}{D}, \amb{E}{P}, and \amb{A}{y}, the proposed model generates ambigrams somewhat similar to human experts. However, especially \amb{C}{D} and \amb{E}{P}, we can find different types of ambigrams between our model and humans. These facts imply that human experts and our models can work complementary to design readable and diverse ambigrams.
\par
It is interesting to note that the class pairs are often limited in the ambigrams by human experts. For example, \amb{B}{Q} is not prepared, and therefore, we need to use \amb{b}{q} instead. (Recall the discussion on interchangeability between capitals and small letters in Section~\ref{sec:quantitative}.) This might be because the design of \amb{B}{Q} was not straightforward  for human experts. In contrast, for our model, the ambigrams of \amb{B}{Q} and \amb{B}{q} are rather easy because these ambigramability scores are 74 and 76, respectively. Fig.~\ref{fig:05-00}, as well as Fig.~\ref{fig:compare-with-manmade}, shows the examples of \amb{B}{Q} generated by the proposed model. A similar observation can be done for \amb{h}{q}, which is difficult for human experts (and therefore only \amb{H}{Q} is designed) but not for our model.\par
\subsubsection{Comparison with Ambigrams by GAN}
To the authors' best knowledge, there has been no past attempt to generate ambigrams by computers. Therefore, there are neither baselines nor state-of-the-art methods which are comparable with our model. Considering this fact, we prepare our own comparative model, using GAN instead of DDPM, to validate the proposed model.\par
%
Fig.~\ref{fig:ambigram-gan} shows the GAN-based model. This model consists of one image generator and two discriminators. The generator is trained (i.e., updated) especially when the discriminator does not determine the generated image as a real sample of class $c$ and/or the 180$^\circ$-rotated version as $c'$. The discriminator is trained when it cannot discriminate a generated image and its rotated image from a real image of a class $c$ and $c'$. The model structures of the generator and discriminator are the same as StyleGAN2~\cite{karras2020analyzing}. This GAN-based model also can be trained end-to-end.\par
%
In this comparative study, we use Google Fonts\footnote{{\tt https://fonts.google.com/}} instead of the MyFonts dataset. This is because the GAN-based model trained by the MyFonts dataset tends to generate hardly-readable noisy images. The MyFonts dataset contains various fonts, including heavily decorative fonts. These variations are beneficial to train the model for the ambigrams generation. However, GAN (especially its generator) cannot deal with the scattered distribution due to the variations. In contrast, Google Fonts contains about 3,000 fonts, most of which have standard shapes and higher legibility. The GAN-based model could perform much better with Google Fonts datasets than the MyFont. Another treatment is to limit the number of classes to 26 (i.e., \letter{A}-\letter{Z}) instead of 52. This limitation also helped the GAN-based model.\par
%
Fig.~\ref{fig:12-50} shows the ambigramability scores evaluated by the same classifier. The GAN-based model has very low scores for many pairs. In contrast, the proposed model achieves higher scores for many pairs.
These results show that the proposed model can generate successful ambigrams more than the GAN-based model.\par
%
Fig.~\ref{fig:12-00} shows the ambigrams generated by (a)~the GAN-based model and (b)~the proposed model. As indicated by the ambigramability scores of Fig.~\ref{fig:12-50}, the ambigrams by the GAN-based model are often inferior to the proposed model. Moreover, it is more prominent that the generated ambigrams by the GAN-based model show fewer variations. 
In \amb{E}{I} and \amb{B}{Y}, this limited variation results in a higher ambigramability score than the proposed model. Of course, it does not mean the success of the GAN-based model. \par
%
This is similar to a situation called {\em mode collapse}, which is specific to GAN. In the standard (i.e., non-ambigram) letter generation scenario, we confirmed that the same generator and discriminator could generate letter images in sufficient style variations. (Again, they are the models used in StyleGAN2, which is famous for good image generation performance.) However, if we use them in the framework of Fig.~\ref{fig:ambigram-gan} for ambigram generation, variations in the generated images decrease drastically. This might be because it is difficult for the generator to find the narrow distributions of ambigrams, which must satisfy dual readability. In other words, DDPM could avoid the mode collapse problem and thus generate not only readable but also diverse images. 

\section{Conclusion}
In this paper, we tackled a novel image generation task, i.e., ambigram generation, and proposed a diffusion model.
The proposed model generates high-quality and diverse ambigrams. For example, if we specify two letter classes, \letter{B} and \letter{a}, as conditions, the proposed model generates ambigrams that can be read as \letter{B} in their original direction and as \letter{a} from a direction rotated $180^\circ$ degrees. An important property of the proposed model is that it does not require any reference ambigrams; it generates ambigrams only using the readability of standard letter images.\par
%
Qualitative and quantitative evaluations show that the proposed model can generate ambigrams with sufficient readabilities and variations for many class pairs. At the same time, we also reveal that the ambigrams for specific class pairs, such as \letter{D} and \letter{K}, are difficult. We  proposed {\em ambigramability} as a metric to automatically and objectively evaluate how easy it is to generate readable ambigrams for individual class pairs.\par
%
A limitation of the current model is that it does not generate word-wise ambigrams. Although they can be realized by arranging the letter-wise ambigrams generated by the current model, it would be more efficient to generate word-wise ambigrams at once. Applying the proposed model to non-Latin alphabets will be another future work.
%
%
%
%
\bibliographystyle{splncs04}
\bibliography{main.bib}


\end{document}